\documentclass[
twocolumn,
]{ceurart}

\usepackage{listings}
\usepackage{subcaption}

\begin{document}

\copyrightyear{2022}
\copyrightclause{Copyright for this paper by its authors. Use permitted under Creative Commons License Attribution 4.0 International (CC BY 4.0).}

\conference{
The IJCAI-ECAI-22 Workshop on Artificial Intelligence Safety (AISafety 2022),
July 24-25, 
2022, 
Vienna, 
Austria}

\title{Privacy Safe Representation Learning via Frequency Filtering Encoder}

\author{Jonghu Jeong}[%
email=jonghu.jeong@deepingsource.io,
]
\author{Minyong Cho}[%
email=minyong.cho@deepingsource.io,
]
\author{Philipp Benz}[%
email=philipp.benz@deepingsource.io,
]
\author{Jinwoo Hwang}[%
email=jinwoo.hwang@deepingsource.io,
]
\author{Jeewook Kim}[%
email=jeewook.kim@deepingsource.io,
]
\author{Seungkwan Lee}[%
email=seungkwan.lee@deepingsource.io,
]
\author{Tae-hoon Kim}[%
email=pete.kim@deepingsource.io,
]
\address[]{Deeping Source Inc.,
  508, Eonju-ro, Gangnam-gu, Seoul, Republic of Korea}

\begin{abstract}
  Deep learning models are increasingly deployed in real-world applications. These models are often deployed on the server-side and receive user data in an information-rich representation to solve a specific task, such as image classification. Since images can contain sensitive information, which users might not be willing to share, privacy protection becomes increasingly important. Adversarial Representation Learning (ARL) is a common approach to train an encoder that runs on the client-side and obfuscates an image. It is assumed, that the obfuscated image can safely be transmitted and used for the task on the server without privacy concerns. However, in this work, we find that training a reconstruction attacker can successfully recover the original image of existing ARL methods. To this end, we introduce a novel ARL method enhanced through low-pass filtering, limiting the available information amount to be encoded in the frequency domain. Our experimental results reveal that our approach withstands reconstruction attacks while outperforming previous state-of-the-art methods regarding the privacy-utility trade-off. We further conduct a user study to qualitatively assess our defense of the reconstruction attack.
\end{abstract}


\maketitle

\section{Introduction}
\begin{figure}[t]
\centering
\includegraphics[width=1.0\linewidth]{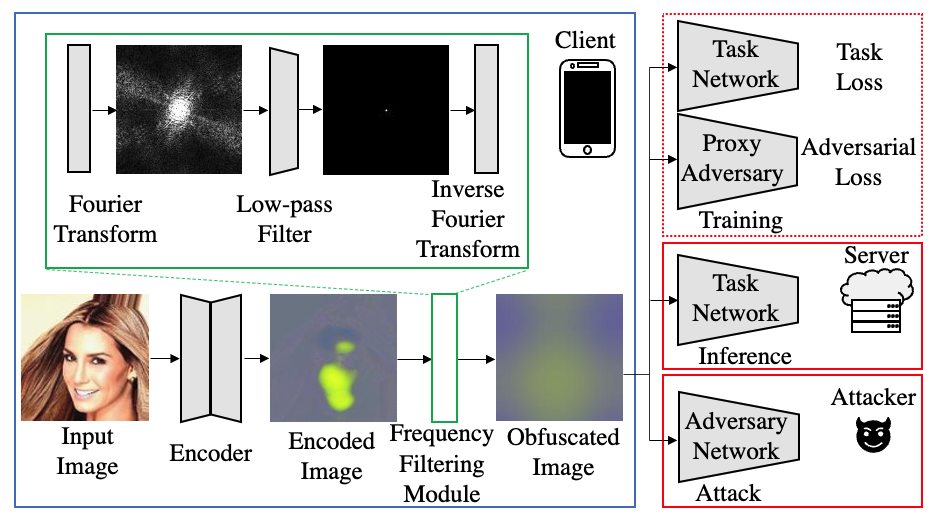}
\caption{An overview of our proposed method. The proposed method trains an encoder that obfuscates an input image through a neural net and leverages a frequency filtering module to safely transmit a privacy-sensitive image from a client-side to a server-side. The frequency filtering module helps the encoder to remove private information effectively from the image while retaining its utility to be used for a particular task of interest (\textit{utility task}) on the server-side. The encoder is trained with the conventional ARL scheme and then deployed to the client-side. Even with the possibility of data leakage during data transmission, malicious attackers can not abuse the obfuscated image for a privacy breach attack (\textit{privacy task}) since the transmitted data contains information that is only useful for the utility task.
}
\label{fig:intro_figure_main_method}
\end{figure}

Service providers, such as Amazon Rekognition and Microsoft Cognitive Services, frequently deploy deep learning models in real-world applications in recent years. The models run on the providers' server can receive and process user information in an information-rich representation to solve a specific task. For example, the users send their face images from their smartphone (client) to the server and receive the processed results, such as face identification. However, the raw images can also contain additional information which users do not consent to reveal or share, violating the users' privacy. An adversary could take over and abuse the images of the users. In one possible attack scenario, adversaries can train a new attacker model (\textit{e.g.} neural network) that retrieves private attributes, such as gender, emotional state, and race. Even the service provider could have malicious intent without the users' knowledge. Hence, an obfuscation method should be used to protect the users' privacy.

For privacy protection with deep learning models, several prior works exist ranging from federated learning~\cite{konevcny2016federated,kairouz2021advances}, split learning~\cite{gupta2018distributed,vepakomma2018split}, differential privacy~\cite{dwork2008differential,ji2014differential,abadi2016deep}, and homomorphic encryption~\cite{hesamifard2017cryptodl,juvekar2018gazelle,nandakumar2019towards} to instance hiding mechanisms~\cite{fu2019mixup,huang2020instahide,shin2020xor,borgnia2021dp}, GAN-based obfuscation techniques~\cite{kim2019training,xu2019ganobfuscator} and adversarial representation learning~\cite{donahue2019large-arl}. 
Among these works, however, adversarial representation learning (ARL) is the one suitable for the service provider to serve users with an obfuscation method. 
For example, federated learning and instance hiding focus on model training with privacy-safe data, not on inference with obfuscated data~\cite{konevcny2016federated,fu2019mixup}. Furthermore, several existing methods suffer under privacy leakage~\cite{lyu2020threats,pasquini2021unleashing,li2021label}, and the degree of computational complexity is too large to be deployed in practice~\cite{hesamifard2017cryptodl,juvekar2018gazelle,nandakumar2019towards}. 
With ARL, the service provider can train an obfuscator model and deploy it to make data obfuscation possible on the user side~\cite{bertran2019adversarially,singh2021disco}.

\begin{figure}[t]
\centering
\includegraphics[width=1.0\linewidth]{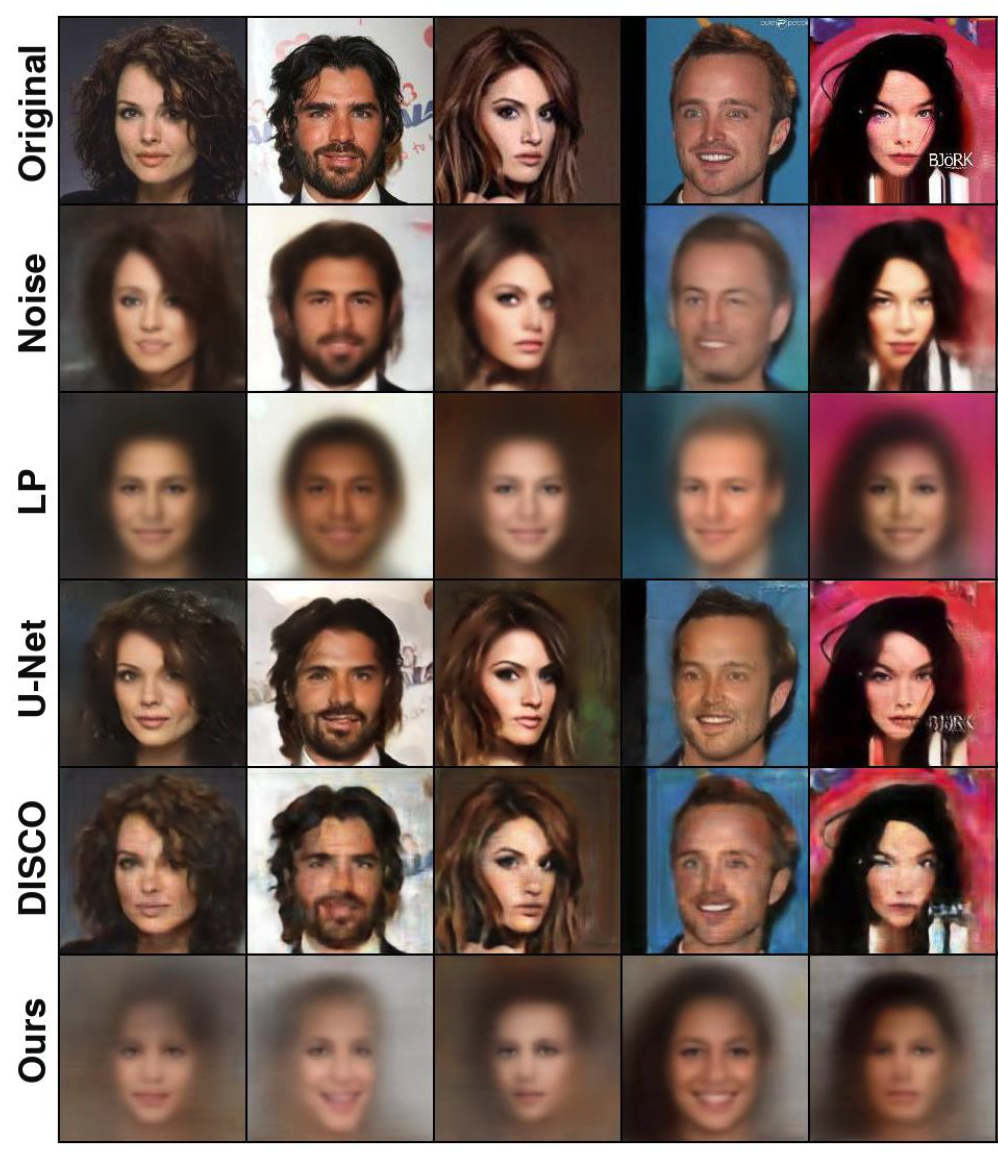}
\caption{
Results of the reconstruction attack with various methods on CelebA. For a successful defense, the reconstructed image should not reveal 1) the identity of the original image and 2) the privacy attribute (in this case, \textit{gender}). Our method successfully defends the reconstruction attack while all other approaches fail. Detailed results are further discussed in Section~\ref{sec:experiments}.
}
\label{fig:recon}
\end{figure}

Most previous ARL methods solve the problem of privacy-safe transmission by optimizing 1) utility task loss and 2) proxy adversary task loss~\cite{roy2019mitigating,bertran2019adversarially,li2021deepobfuscator,singh2021disco}. They also introduce specific loss-design formulations, model architecture design, and training schemes. The methods are evaluated quantitatively with performance on both utility and adversary tasks. Note that there usually exists a trade-off between privacy and utility. We use a reconstruction attack, to test the quality of the obfuscation. In a reconstruction attack, a new model is trained that takes the obfuscated representation as an input and outputs the original image. As demonstrated in Figure~\ref{fig:recon}, the original data of existing ARL methods can successfully be recovered from the obfuscated representation. This result suggests that the private information is still encoded in the obfuscated representations.

We present a novel ARL method that leverages frequency filtering, leveraging an extreme low-pass frequency filter (Figure~\ref{fig:intro_figure_main_method}).
The representation filtering on the frequency domain effectively limits the amount of information to be encoded. Our experimental results show that our approach outperforms previous state-of-the-art methods regarding the privacy-utility trade-off. We also present that our proposed method withstands the reconstruction attack better than existing ARL methods, which are evaluated through visual metrics and a user study.

\section{Related Work}

\paragraph{Data-privacy in Computer Vision}
For privacy-safe data transmission, several approaches have been proposed to tackle the problem of raw image sharing. Federated learning~\cite{konevcny2016federated,kairouz2021advances} and split learning~\cite{gupta2018distributed,vepakomma2018split} aim to train a machine learning model without directly sharing raw images through sharing gradients or a processed representation. These methods usually focus on the model training, and not on inference with obfuscated data.
Homomorphic encryption~\cite{hesamifard2017cryptodl,juvekar2018gazelle,nandakumar2019towards} attempts to train models on encrypted data, such that the data can be shared in encrypted form and be processed without decryption. Currently, this method suffers from a considerably high computational cost. Instance hiding mechanisms~\cite{fu2019mixup,huang2020instahide,shin2020xor,borgnia2021dp} introduce random pixel mixing and clipping algorithm to perturb images. The perturbed images are used only for the training, and the original images are used for the inference
which means that there are still potential threats for data breaches when inferring the target.

\paragraph{Adversarial Representation Learning (ARL)} 
Another line of work focuses on the training framework of ARL to address the utility-privacy trade-off of (a) mitigation of privacy disclosure while (b) maintaining task utility. ARL methods have found their application in practical scenarios, such as information censoring~\cite{edwards2016censoring}, learning fair representations~\cite{louizos2016variational,madras2018learning}, the mitigation of information leakage~\cite{roy2019mitigating,bertran2019adversarially,li2021deepobfuscator}, collaborative inference~\cite{vepakomma2020nopeek,osia2020hybrid,singh2021disco}, and GAN-based obfuscation techniques~\cite{kim2019training,xu2019ganobfuscator}.
Commonly, the ARL framework consists of three entities:
1) an obfuscator, which transforms input data to a private representation that retains utility,
2) a task model, performing the utility task on the data representation,
3) a proxy adversary, attempting to extract sensitive attributes.
Recent approaches~\cite{pittaluga2019learning,liu2019privacy,wu2018towards,li2021deepobfuscator} represent each component as deep neural networks (DNNs).
MaxEnt~\cite{roy2019mitigating} formulate the ARL problem as an adversarial non-zero-sum game and minimizes the amount of non-utility information, which they quantify through entropy. Adversarial representation learning with non-linear functions through kernel representation with theoretical guarantees are introduced in~\cite{sadeghi2019global}. 
While most of the previous methods represent the obfuscated output as the intermediate feature of a DNN,
\citeauthor{bertran2019adversarially}~\cite{bertran2019adversarially} leverages domain-preserving transformations, \textit{i.e.} images to images.
Above mentioned ARL methods mainly focused on designing special loss functions or model architectures.
To the best of our knowledge, our method is the first ARL method that focuses on the effective encoding of privacy-safe representation in the frequency domain.

There are three common attacks on privacy in machine learning.
The first is the membership inference attack~\cite{shokri2017membership}, which attempts to infer whether a data sample is used for the machine learning model training. This attack is more related to the attack on the server-side model, not the transmitted data.
The second is the inversion attack~\cite{fredrikson2015model} which attempts to infer raw data from processed representation. This is the same attack scenario as the aforementioned reconstruction attack. 
The last is the information leakage attack~\cite{roy2019mitigating}, for which adversaries attempt to infer privacy-related information from obfuscated representation.
In this work the inversion attack and the information leakage attack are considered as they are potential threats to transmitted privacy-sensitive images.

\paragraph{Frequency Perspective in Computer Vision}
Prior works have explored the behavior of DNNs from a frequency perspective.
Overall, there is solid evidence that both high-frequency features and low-frequency features can be helpful for classification~\cite{yin2019fourier,wang2020high}.
It has been demonstrated that DNNs have an increased bias toward texture compared to the object's shape~\cite{geirhos2018imagenet}.
On the other hand, DNNs trained only on low-pass filtered images also generalize well and are capable of achieving high accuracies~\cite{yin2019fourier}.
\citeauthor{yin2019fourier}~\cite{yin2019fourier} shows that adversarial training and Gaussian data augmentation shift DNNs towards utilizing low-frequency information in the input.
\citeauthor{wang2020high}~\cite{wang2020high} points out that convolutional neural networks (CNNs) mainly exploit high-frequency components.
Similarly,~\citeauthor{abello2021dissecting}~\cite{abello2021dissecting} find that mid or high-level frequencies are disproportionately critical for CNNs.
\citeauthor{ilyas2019adversarial}~\cite{ilyas2019adversarial} also show similar findings that human-imperceptible features with high-frequency properties are sufficient for the model to exhibit high generalization capability. 

In this work, we leverage previous insights that information can be encoded in different frequency ranges of images. We propose encoding information in the low-frequency band of images to securely transfer them between different parties.

\section{Problem Formulation}
We consider an image dataset $x \sim \mathcal{X} \in \mathbf{R}^{H \times W \times 3}$, where $H$ and $W$ represent width and height, respectively, along with a number of various attributes $y \sim \mathcal{Y}$. Some of the attributes are private attributes $y_p \sim \mathcal{Y}_p$ and some are utility attributes $y_t \sim \mathcal{Y}_t$, such that $\mathcal{Y} = \mathcal{Y}_t \cup \mathcal{Y}_p$. Given a utility task model $f_t$, we search for an intermediate representation $\hat{x}$, from which $f_t$ can infer the utility attributes, but not the privacy attributes. This transformation can also be represented through a DNN $o$, termed obfuscator, resulting in $o(x) = \hat{x}$. Note that in prior works, the intermediate representation $\hat{x}$ was often represented as a feature map differing in shape from the original input images. However, similar to~\cite{bertran2019adversarially}, we represent the obfuscated representation in the same shape as the original input image. This setting allows us to leverage existing image transformation techniques, such as transforming them into a 2D Fourier representation. Additionally, this form of intermediate representation allows us to analyze the representations visually.

\paragraph{Threat Model}
Given the above problem formulation, an attacker can attempt to retrieve information about the private attributes from the intermediate representation. This can be realized either by directly inferring private information from the intermediate representation (\textit{information leakage attack}) or through the reconstruction of the original input images from the intermediate representations (\textit{reconstruction attack}). In the \textit{information leakage attack} scenario an attacker is able to obtain data pairs consisting of the corresponding intermediate representation and their respective private attributes $\{ \hat{x}, y_p \}$. In this scenario an attacker can attempt to train a model $f_a$, which leaks the private information from the representations $f_a(\hat{x}) = y_p$. In the \textit{reconstruction attack}, given image pairs of the original image and the intermediate representation $\{ x, \hat{x} \}$ the attacker attempts to obtain a model $f_r$ which retrieves the original image $x$ from the intermediate representation $f_r(\hat{x}) = x$. In this work, we represent both attacker models $f_a$ and $f_r$ through DNNs, since they are proven to be powerful for image processing tasks.

\section{Methodology}
\paragraph{Fourier Transformation}
Fourier transform is a common tool to perform frequency analysis~\cite{lim1990two}. We consider the 2D discrete Fourier transformation $\mathcal{F} : \mathbf{R}^{W\times H} \rightarrow \mathbf{C}^{W \times H}$ and the inverse Fourier transformation as $\mathcal{F}^{-1}$. After applying $\mathcal{F}$ on an image, low frequencies are located in the center of a Fourier image, while high frequencies are located toward the boundaries.
For low-pass filtering, we set all frequency components outside of a central circle with radius $r$ in the frequency domain to zero and apply $\mathcal{F}^{-1}$ afterward. We normalize the radius to be in the range of [0, 1] by considering the center of the image as 0 and the corner as 1. We indicate low-pass filtering as $LP$.

\paragraph{Frequency Obfuscation}
We depict our proposed methodology in Figure~\ref{fig:intro_figure_main_method}. Given an input image, the objective is to obfuscate the image to achieve the best privacy-utility trade-off. Our obfuscator module consists of an encoder architecture followed by frequency-filtering. We choose the commonly used U-Net~\cite{ronneberger2015u} architecture as our encoder and pass the original image through it. Formally, we express this as $e(x)$, where we indicated the encoder with $e$. The subsequent frequency filtering is realized via a low-pass filter $LP(e(x))$. This procedure completes the generation of the intermediate representation through the obfuscator $\hat{x} = o(x) = LP(e(x))$. During obfuscator training, we leverage a task model and a proxy adversary. 
The objective of the task model is to predict the utility attribute from the intermediate representation. The respective task loss can be calculated with $l_t=\mathbb{E}[\mathcal{L}_t(f_t(o(x)), y_t)]$, where $\mathcal{L}_t$ indicates the task loss function, which is the cross-entropy function in our setup. 
The objective of proxy adversary model is to leak the privacy attribute from the intermediate representation. The proxy adversary loss can be calculated as $l_p=\mathbb{E}[\mathcal{L}_p(f_a(o(x)), y_p)]$, where $\mathcal{L}_p$ indicates the privacy loss function, which is also represented as the cross-entropy function. The obfuscator loss is represented as $l_o = l_t - l_p$.

Similar to the scenario introduced in DISCO~\cite{singh2021disco} a practical application scenario of our proposed approach is when the obfuscator module is present on a trusted client device, which sends the intermediate feature representations to a server. Since an adversary can intercept the communication between client and server, or the server can also be malicious, we consider the server-side an untrusted entity.

\paragraph{Evaluation Protocol} 
In the following, we outline our evaluation protocol.
We follow the general ARL evaluation protocol~\cite{singh2021disco,roy2019mitigating}. 
Given an image classification dataset, we specify certain classes as the utility and privacy tasks, respectively. Based on the chosen tasks, following our proposed method we obtain an obfuscator and a utility task model. Note that this includes training proxy adversaries. After training, we evaluate the models on the utility task and report the accuracy as \textit{utility}. Then we freeze the weights of the obfuscator and train an adversary model to predict the privacy attributes and report the accuracy as \textit{privacy}. To assess the privacy-utility trade-off, we measure their difference ($\Delta$).

Additionally, we report the \textit{performance bounds}. Theoretically, the utility (higher the better) is upper bounded by $100\%$. In practice, however, we consider the upper bound as the utility performance of a ResNet18~\cite{he2016deep} model trained on the original images.
For privacy (lower the better), we consider the lower bound as the random guess for the privacy attribute.

We also perform a reconstruction attack on the obfuscated images to recover corresponding original images. We evaluate the reconstruction attacks quantitative and qualitatively by calculating similarity scores between the original and reconstructed images and conducting a user study on the reconstructed images. 

\section{Experiments}
\label{sec:experiments}

\begin{table*}[ht!]
\setlength\tabcolsep{3.0pt}
\centering
\scalebox{1.2}{
\begin{tabular}{l | ccc | ccc | ccc}
\toprule
      & \multicolumn{3}{c}{Fairface} & \multicolumn{3}{c}{CelebA} & \multicolumn{3}{c}{CIFAR10} \\
Method & Privacy $\downarrow$ & Utility $\uparrow$ & $\Delta \uparrow$ & Privacy $\downarrow$ & Utility $\uparrow$ & $\Delta \uparrow$ & Privacy $\downarrow$ & Utility $\uparrow$ & $\Delta \uparrow$ \\
\midrule
Perf.\ Bounds & 19.03 & 90.16 & 71.13 & 57.43 & 93.32 & 35.89 & 10.00 & 98.79 & 78.79 \\
\midrule
Noise & 42.61 & 74.33 & 31.72 & 91.71 & 85.38 & -6.33 & 54.37 & 87.77 & 33.40\\
LP    & 31.93 & 64.77 & 32.84 & 76.52 & 63.69 & -12.83 & 47.05 & 85.76 & 38.71\\
U-Net & 51.52 & 86.40 & 34.88 & 87.21 & 93.12 & 5.91 & 85.05 & 95.45 & 10.40\\
DISCO & 19.00 & 81.50 & 62.50 & 61.20 & 91.00 & 29.80 & 22.30 & 91.98 & 69.68 \\
Ours  & 23.63 & 89.67 & \textbf{66.04} & 61.60 & 93.27 & \textbf{31.67} & 22.58 & 92.95 & \textbf{70.37}\\
\bottomrule
\end{tabular}
}
\caption{Evaluation of the privacy-utility trade-off. The upper/lower arrow suggests that each value is higher/lower the better. Our method shows the biggest gap between privacy and utility accuracy among all the datasets. Note that the privacy accuracy is based on the newly trained adversary model which is trained with the fully trained and frozen obfuscation model.} 
\label{tab:sensitive-attribute}
\end{table*}

\subsection{Setup}
\label{sec:experiments_setup}
\paragraph{Datasets}
We conduct experiments on CelebA~\cite{liu2015faceattributes}, FairFace~\cite{karkkainen2021fairface}, and CIFAR10~\cite{Krizhevsky09learningmultiple}. Following the utility and privacy task setting from DISCO~\cite{singh2021disco}, we set ``Smiling" as the utility attribute and ``Male" as the privacy attribute for CelebA, ``Gender" as the utility attribute, and ``Race" as the privacy attribute for FairFace. For CIFAR10, the utility task is defined as classifying living objects (\textit{e.g.} ``bird", ``cat", \textit{etc.}) or non-living objects (\textit{e.g.} ``airplane", ``automobile", \textit{etc.}) and the privacy task as classifying the separate 10 classes.

\paragraph{Implementation details}
The encoder is a lightweight variant of U-Net~\cite{ronneberger2015u}, with 4$\times$ fewer intermediate feature channels than the original version.
We use an extreme low pass filter with radius, $r=0.01$ for CelebA and FairFace, and $r=0.05$ for CIFAR10.
We apply a center-circled filter, which can adjust the level of obfuscation by changing its radius (bandwidth). Section~\ref{sec:ablation_radius} discusses the effect of the radius. We normalize the radius by the length from the filter's center to the corner to make the value in the range $[0, 1]$. For both the utility and privacy task models, we use ResNet-18~\cite{he2016deep}, and use the same dataset for training both models. We use Adam~\cite{kingma2014adam} optimizer for all 3 models with learning rate $10^{-4}$ for U-Net and $10^{-3}$ for the ResNet-18 models. We evaluate the top-1 accuracy for both utility and privacy tasks. 
We used the lightweight U-Net as the reconstructor for the reconstruction attack. The reconstructor adversary is trained with the MSE loss between the original and the reconstructed images. The reconstructed images are evaluated using MSE, $L_1$, SSIM~\cite{zhou2004ssim}, MS-SSIM~\cite{wang2003multiscale-msssim}, PSNR~\cite{hore2010psnr}, and LPIPS~\cite{zhang2018unreasonable-lpips}. MSE, $L_1$, and PSNR compare the images pixel-wise while SSIM and MS-SSIM compare structural similarity (\textit{e.g.}, brightness, contrast) between the images. LPIPS uses a pre-trained neural network's feature map for comparison. These metrics are commonly used for comparing the similarity between images~\cite{singh2021disco,li2021deepobfuscator,Karras_2020_CVPR} and we consider them as a proxy of human vision.

\paragraph{Compared Methods}
We compare our method with various baselines. As a simple baseline obfuscator, we add Gaussian noise sampled from $\mathcal{N}(0,\,\sigma^{2})$ to the input image while obeying the image range of pixels in the range $[0,1]$. We indicate this method with \textit{Noise}.
We use $\sigma^{2}=4$ for CelebA and FairFace and $\sigma^{2}=0.64$ for CIFAR10, which obfuscate the images sufficiently.
To investigate the sole effect of the low-pass filtering, we apply only the low-pass filter to the raw images. We name this baseline as \textit{LP}. Complementary, we also compare the U-Net without the low-pass filtering module as an obfuscator. We call it \textit{U-Net}. This setup is similar to DeepObfuscator~\cite{li2021deepobfuscator} which uses an encoder, task model, and a proxy adversary. However, since DeepObfuscator has not open-sourced their code, we used our U-Net encoder as a method to compare. Finally, we compare our method to the state-of-the-art ARL method \textit{DISCO}~\cite{singh2021disco}, which selectively removes features via channel pruning in the latent space.

\subsection{Results}
\label{sec:results}

Table~\ref{tab:sensitive-attribute} shows a comparison between the privacy and utility accuracy of each obfuscation method. Our method resulted in the highest gap between utility and privacy accuracy on all datasets. For the methods without encoder (\textit{i.e.} \textit{Noise} and \textit{LP}), the accuracy for both utility and privacy decreases compared to training with the original image since these methods obfuscate images without any prior knowledge of the tasks. These methods cannot selectively restrict information for high utility and low privacy leakage. \textit{U-Net} showed high utility accuracy but failed to defend against the privacy attack, although it is trained with a proxy adversary. We conjecture that simply taking the guidance of the proxy model loss is not enough for the encoder to learn to restrict information. Our method is a combination of \textit{LP} and \textit{U-Net}, and learns to encode a representation into the restricted bandwidth, which is limited by the frequency filtering module. This limited bandwidth helps the encoder to learn how to extract utility information effectively and remove privacy attributes to fully leverage the limited bandwidth. While the same data is used to train both utility and adversary models, which is a generous and unrealistic condition for the attackers to have, we found the adversary model performed poorly. \textit{DISCO} shows the lowest privacy accuracy among all the datasets. However, the utility accuracy is lower than our method, so the utility-privacy gap is smaller than ours. 

\begin{table}
\setlength\tabcolsep{1.0pt}
\centering
\begin{tabular}{l | r|r|r|c|c|c }
\toprule
Method & MSE $\uparrow$ & $L_1$ $\uparrow$ & SSIM $\downarrow$ & MS-SSIM $\downarrow$ & PSNR $\downarrow$ & LPIPS $\uparrow$ \\
\midrule
Noise & 584.88 & 16.97 & 0.6017 & 0.7776 & 20.46 & 0.3714 \\
LP & 1889.15 & 32.10 & 0.4632 & 0.5390 & 15.37 & 0.5537 \\
U-Net & 390.34 & 13.81 & 0.7505 & 0.8839 & 22.22 & 0.1809 \\
DISCO & 567.17 & 15.94 & 0.5765 & 0.7611 & 20.60 & 0.4351 \\
Ours & \textbf{3689.50} & \textbf{48.08} & \textbf{0.4240} & \textbf{0.4728} & \textbf{12.47} & \textbf{0.6145} \\
\bottomrule
\end{tabular}
\caption{Similarity scores between the original image and the reconstructed ones on CelebA. The upper/lower arrow suggests that each value is higher/lower the better, respectively. Our approach shows the best dissimilarity among all the metrics.}
\label{tab:sim_score}
\end{table}

In terms of the visual quality, our obfuscated representations appear as simple globs of color, making them unrecognizable to human observers (Figure~\ref{fig:intro_figure_main_method}). The obfuscated representations from other methods also appear obfuscated to the human eye. However, applying our best effort reconstruction attack, it is possible to reconstruct the original image or infer the privacy attribute (\textit{i.e.} gender) from reconstructed images. (Figure~\ref{fig:recon}).
The reconstructed images from our method successfully defend identity reconstruction and privacy attribute leakage, with the reconstructed images all being relatively similar to each other. The quantitative results of the reconstruction attack in Table~\ref{tab:sim_score} further confirm this since all scores achieve the best results in terms of dissimilarity for our approach.
We note that an adversary model trained with the reconstructed images to infer the privacy attributes performs worse than directly training the model with the obfuscated images since the reconstructed images are processed from the obfuscated images. 

\subsection{User Study}
\label{sec:user_study}

\begin{figure}[t]
\centering
\includegraphics[width=1.0\linewidth]{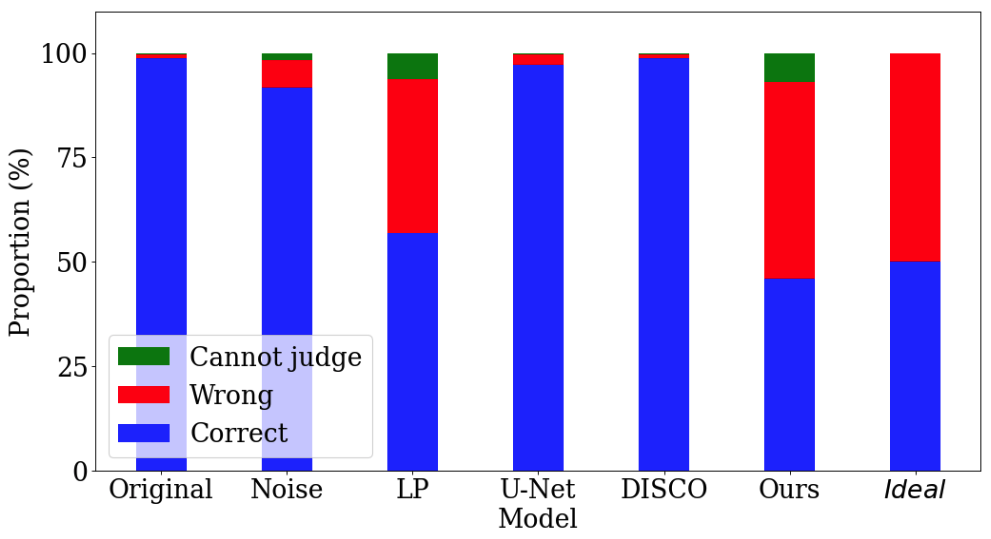}
\caption{
Result of the user study on reconstructed images of CelebA. We asked the participants to classify \textit{gender} (male/female) on 180 images such as Figure~\ref{fig:recon}. The participants correctly distinguished the gender of original images and reconstructed images from the three methods (\textit{Noise}, \textit{U-Net}, and \textit{DISCO}) with more than 90\% accuracy. Our method and \textit{LP} effectively confused the participants with gender-neutral faces (45.83\% and 56.9\% of correct answers ratio each), while ours is slightly better than \textit{LP} in terms of obfuscation. We also plot the ideal case of the user study to show our method's near-perfect superiority against the reconstruction attack.
}
\label{fig:user_study}
\end{figure}

We present a user study to show our method’s robustness against the reconstruction attack on CelebA. Since the privacy task for the dataset is gender classification, the reconstructed image’s gender should not be correctly classified by a human observer if the obfuscation is successful. To conduct the experiment, we randomly sampled 30 images (15 for male and 15 for female), for which ResNet18 classifies the gender correctly. By doing so, we balanced each class and addressed the ambiguity of the labels to prevent unfair results.
Then, we obfuscated the images using each of the techniques and reconstructed them with their respective attacker models from Section~\ref{sec:experiments_setup}. Examples of reconstructed images are shown in Figure~\ref{fig:recon}.
We presented 180 reconstructed images to a group of people and asked them to identify whether the person in the reconstructed image is male, female, or cannot be judged. We provided the last option to let the users skip the examples that are hard to judge. The test subjects were randomly selected and consist of 30 people who live in Seoul, South Korea, and are in their 20s and 30s.

As shown in Figure~\ref{fig:user_study}, people correctly identify the gender for the original images and the reconstructed ones from the methods \textit{Noise}, \textit{U-Net}, and \textit{DISCO}. 
More than 90\% of answers were correct for the three methods.
\textit{LP} showed a relatively low correct ratio (56.9\%) and a high ``cannot judge" ratio (6.19\%). Our method showed the best for both, the lowest correct ratio of 45.83\% and the highest ``cannot judge" ratio of 7.02\%. 
We consider the 50\% ratio for each ``correct" and ``wrong" answer as a random guess since the labels for the test datasets are balanced. Additionally, we note that ``cannot judge" can be considered as a random guess since without this option, the users would have done a random choice. The results indicate that our approach successfully protects against reconstruction attacks in terms of human vision. The results also align with the quantitative results (Table~\ref{tab:sim_score}). In terms of obfuscation, our method shows the best results, followed by \textit{LP}. It reconfirms the usefulness of our architecture design, the combination of the encoder and the frequency filtering module.

\section{Ablation Study}
\label{sec:ablation}

\subsection{High-pass filter}
\label{sec:ablation_high_freq}

\begin{table}
\centering
\begin{tabular}{l | ccc }
\toprule
Method & Privacy $\downarrow$ & Utility $\uparrow$ & $\Delta \uparrow$  \\
\midrule
\textit{HP} (r=0.80) & 26.19 & 89.03 & 62.84 \\
\textit{HP} (r=0.85) & 26.28 & 89.13 & 62.85 \\
\textit{HP} (r=0.90) & 28.94 & 88.00 & 59.06 \\
\textit{HP} (r=0.95) & 24.96 & 88.12 & 63.16 \\
\textit{HP} (r=0.99) & 19.03 & 52.88 & 33.85 \\
\midrule
\textit{LP} (r=0.01) & 23.63 & 89.67 & \textbf{66.04}  \\
\bottomrule
\end{tabular}
\caption{The privacy-utility gap of the high-pass filtering module on FairFace. Our low-pass filtering module shows the best privacy-utility gap compared to the high-pass filter with the various filter radii.
}
\label{tab:high_freq_table}
\end{table}

Previously, we presented the effect of the low-pass frequency filtering module on ARL. The module appropriately limits the amount of encoded information in the obfuscated image. It retains the information at a low-frequency range. Using a high-pass filter, we can leverage the same intuition, by limiting the information to be encoded in the high-frequency bandwidth. However, in the following, we will present results indicating that the low-pass filter is the superior method to use.

We conduct the same experiment from Section~\ref{sec:results} on FairFace with a high-pass filtering module for 5 radii (0.80, 0.85, 0.90, 0.95, 0.99). Contrary to the low-pass filtering, the filter removes frequencies inside the filter radius, which leads to a radius of 0.99 as the most extreme high-pass filter. We call this method \textit{HP}.

The respective results are presented in Table~\ref{tab:high_freq_table}. As the filtering gets more extreme, the utility accuracy decreases together with the privacy accuracy. The table also shows that our approach with a low-pass filter from Table~\ref{tab:sensitive-attribute} outperforms all results from the high-pass filter regarding the privacy-utility gap. The best privacy-utility gap with the high-pass filter is 63.16\% with a radius of 0.95, which is 2.88\%p lower than for the approach with low-pass filtering.
It has been demonstrated that DNNs can learn from low-pass filtered images more efficiently than high-pass filtered ones~\cite{yin2019fourier}.
Especially with the extreme high-pass (r=0.99), the model did not learn for both, the utility and privacy tasks.

Furthermore, from a practical point of view, we need to reduce the size of the obfuscated image to reduce the cost of transmission or storage. The most commonly used JPEG compression algorithm leverages the filtering of high frequency. If we use a high-pass filter ARL method, encoded information in the high-frequency range would be lost. To this end, encoding information into the low-frequency range is more suitable than the opposite to utilize the conventional compression algorithms further.

\subsection{The effect of filter radius}
\label{sec:ablation_radius}

\begin{figure}[t]
    \centering
    \begin{subfigure}[b]{0.25\textwidth}
             \centering
             \includegraphics[width=\textwidth]{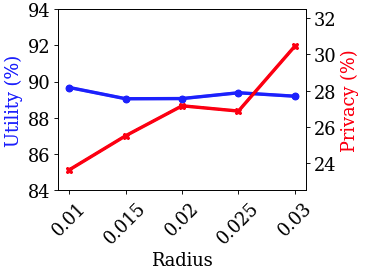}
    \end{subfigure}
    \hfill
    \begin{subfigure}[b]{0.22\textwidth}
             \centering
             \includegraphics[width=\textwidth]{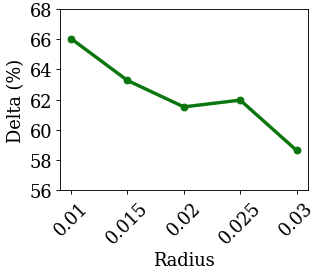}
    \end{subfigure}
    \caption{(Left) Privacy and utility accuracy under each radius of the low-pass filter. The experiments are conducted on FairFace. (Right) Privacy-utility trade-off. Delta represents the performance gap between utility and privacy.}
    \label{fig:low_pass_tradeoff}
\end{figure}

One of the key points of our proposed method is the frequency filtering module. The module has only one parameter to consider, the filter's radius. To gain insight into choosing the parameter, we conducted experiments with various radii. The same experiment from Section~\ref{sec:experiments} on FairFace is done with 5 radii (0.01, 0.015, 0.02, 0.025, 0.03). The radius of 0.01 is the most extreme low-pass filter.

Figure~\ref{fig:low_pass_tradeoff} (left) shows a trend of consistent utility accuracy and increasing privacy accuracy. The utility accuracies are around 89\% with a small variance.
The privacy accuracies show an increasing tendency from 23.64\% to 30.45\% as the radius increases.
It leads the privacy-utility gap to decrease (Figure~\ref{fig:low_pass_tradeoff}, right).

The increased privacy accuracy aligns with our intuition of limiting information in the obfuscated representation. The wider radius allows the representation to have more information, leading the adversary to exploit it for a privacy attack easily.
Note that the utility accuracy did not decrease even with the harshest filter. We speculate that the extremely low-pass filtered representation is enough for these specific utility tasks. 
Figure~\ref{fig:low_pass_tradeoff} and Table~\ref{tab:high_freq_table} confirm that the radius is a crucial factor of privacy and utility accuracy.
Thus the radius is a hyperparameter that should be tuned based on the privacy-utility gap.

\section{Conclusion}

This work proposes a novel ARL method based on frequency filtering, which is robust to privacy leakage attacks while maintaining task utility. Our experiments suggest that a combination of neural-net encoder and low-pass filter improves ARL training for the quantitative and qualitative metrics. The method outperforms other compared methods for the quantitative measure of privacy-utility trade-off and reconstruction attack (Section~\ref{sec:experiments}). Our user study suggests that the proposed method effectively defends against reconstruction attacks (Section~\ref{sec:user_study}). The ablation experiments justified the use of a low-pass filter and also showed that the filter radius adjusts the privacy-utility trade-off (Section~\ref{sec:ablation}). 

For future work we consider the optimization of the client-side model to reduce the computation burden by using a lightweight architecture such as MobileNetV3~\cite{howard2019searching-mobilenetv3}. Furthermore, an adaptive selection of the frequency-filtering hyperparameter might increase the utility accuracy and decrease the privacy accuracy.

\bibliography{bibliography}

\begin{thebibliography}{53}
\expandafter\ifx\csname natexlab\endcsname\relax\def\natexlab#1{#1}\fi
\providecommand{\url}[1]{\texttt{#1}}
\providecommand{\href}[2]{#2}
\providecommand{\path}[1]{#1}
\providecommand{\DOIprefix}{doi:}
\providecommand{\ArXivprefix}{arXiv:}
\providecommand{\URLprefix}{URL: }
\providecommand{\Pubmedprefix}{pmid:}
\providecommand{\doi}[1]{\href{http://dx.doi.org/#1}{\path{#1}}}
\providecommand{\Pubmed}[1]{\href{pmid:#1}{\path{#1}}}
\providecommand{\bibinfo}[2]{#2}
\ifx\xfnm\relax \def\xfnm[#1]{\unskip,\space#1}\fi
\bibitem[{Kone{\v{c}}n{\`y} et~al.(2016)Kone{\v{c}}n{\`y}, McMahan, Yu,
  Richt{\'a}rik, Suresh, and Bacon}]{konevcny2016federated}
\bibinfo{author}{J.~Kone{\v{c}}n{\`y}}, \bibinfo{author}{H.~B. McMahan},
  \bibinfo{author}{F.~X. Yu}, \bibinfo{author}{P.~Richt{\'a}rik},
  \bibinfo{author}{A.~T. Suresh}, \bibinfo{author}{D.~Bacon},
\newblock \bibinfo{title}{Federated learning: Strategies for improving
  communication efficiency},
\newblock \bibinfo{journal}{arXiv preprint arXiv:1610.05492}
  (\bibinfo{year}{2016}).
\bibitem[{Kairouz et~al.(2021)Kairouz, McMahan, Avent, Bellet, Bennis, Bhagoji,
  Bonawitz, Charles, Cormode, Cummings et~al.}]{kairouz2021advances}
\bibinfo{author}{P.~Kairouz}, \bibinfo{author}{H.~B. McMahan},
  \bibinfo{author}{B.~Avent}, \bibinfo{author}{A.~Bellet},
  \bibinfo{author}{M.~Bennis}, \bibinfo{author}{A.~N. Bhagoji},
  \bibinfo{author}{K.~Bonawitz}, \bibinfo{author}{Z.~Charles},
  \bibinfo{author}{G.~Cormode}, \bibinfo{author}{R.~Cummings}, et~al.,
\newblock \bibinfo{title}{Advances and open problems in federated learning},
\newblock \bibinfo{journal}{Foundations and Trends{\textregistered} in Machine
  Learning}  (\bibinfo{year}{2021}).
\bibitem[{Gupta and Raskar(2018)}]{gupta2018distributed}
\bibinfo{author}{O.~Gupta}, \bibinfo{author}{R.~Raskar},
\newblock \bibinfo{title}{Distributed learning of deep neural network over
  multiple agents},
\newblock \bibinfo{journal}{Journal of Network and Computer Applications}
  (\bibinfo{year}{2018}).
\bibitem[{Vepakomma et~al.(2018)Vepakomma, Gupta, Swedish, and
  Raskar}]{vepakomma2018split}
\bibinfo{author}{P.~Vepakomma}, \bibinfo{author}{O.~Gupta},
  \bibinfo{author}{T.~Swedish}, \bibinfo{author}{R.~Raskar},
\newblock \bibinfo{title}{Split learning for health: Distributed deep learning
  without sharing raw patient data},
\newblock \bibinfo{journal}{arXiv preprint arXiv:1812.00564}
  (\bibinfo{year}{2018}).
\bibitem[{Dwork(2008)}]{dwork2008differential}
\bibinfo{author}{C.~Dwork},
\newblock \bibinfo{title}{Differential privacy: A survey of results},
\newblock in: \bibinfo{booktitle}{International conference on theory and
  applications of models of computation}, \bibinfo{year}{2008}.
\bibitem[{Ji et~al.(2014)Ji, Lipton, and Elkan}]{ji2014differential}
\bibinfo{author}{Z.~Ji}, \bibinfo{author}{Z.~C. Lipton},
  \bibinfo{author}{C.~Elkan},
\newblock \bibinfo{title}{Differential privacy and machine learning: a survey
  and review},
\newblock \bibinfo{journal}{arXiv preprint arXiv:1412.7584}
  (\bibinfo{year}{2014}).
\bibitem[{Abadi et~al.(2016)Abadi, Chu, Goodfellow, McMahan, Mironov, Talwar,
  and Zhang}]{abadi2016deep}
\bibinfo{author}{M.~Abadi}, \bibinfo{author}{A.~Chu},
  \bibinfo{author}{I.~Goodfellow}, \bibinfo{author}{H.~B. McMahan},
  \bibinfo{author}{I.~Mironov}, \bibinfo{author}{K.~Talwar},
  \bibinfo{author}{L.~Zhang},
\newblock \bibinfo{title}{Deep learning with differential privacy},
\newblock in: \bibinfo{booktitle}{ACM SIGSAC conference on computer and
  communications security}, \bibinfo{year}{2016}.
\bibitem[{Hesamifard et~al.(2017)Hesamifard, Takabi, and
  Ghasemi}]{hesamifard2017cryptodl}
\bibinfo{author}{E.~Hesamifard}, \bibinfo{author}{H.~Takabi},
  \bibinfo{author}{M.~Ghasemi},
\newblock \bibinfo{title}{Cryptodl: Deep neural networks over encrypted data},
\newblock \bibinfo{journal}{arXiv preprint arXiv:1711.05189}
  (\bibinfo{year}{2017}).
\bibitem[{Juvekar et~al.(2018)Juvekar, Vaikuntanathan, and
  Chandrakasan}]{juvekar2018gazelle}
\bibinfo{author}{C.~Juvekar}, \bibinfo{author}{V.~Vaikuntanathan},
  \bibinfo{author}{A.~Chandrakasan},
\newblock \bibinfo{title}{$\{$GAZELLE$\}$: A low latency framework for secure
  neural network inference},
\newblock in: \bibinfo{booktitle}{USENIX Security Symposium},
  \bibinfo{year}{2018}.
\bibitem[{Nandakumar et~al.(2019)Nandakumar, Ratha, Pankanti, and
  Halevi}]{nandakumar2019towards}
\bibinfo{author}{K.~Nandakumar}, \bibinfo{author}{N.~Ratha},
  \bibinfo{author}{S.~Pankanti}, \bibinfo{author}{S.~Halevi},
\newblock \bibinfo{title}{Towards deep neural network training on encrypted
  data},
\newblock in: \bibinfo{booktitle}{Conference on Computer Vision and Pattern
  Recognition Workshops (CVPR-W)}, \bibinfo{year}{2019}.
\bibitem[{Fu et~al.(2019)Fu, Wang, Xu, Mi, and Wang}]{fu2019mixup}
\bibinfo{author}{Y.~Fu}, \bibinfo{author}{H.~Wang}, \bibinfo{author}{K.~Xu},
  \bibinfo{author}{H.~Mi}, \bibinfo{author}{Y.~Wang},
\newblock \bibinfo{title}{Mixup based privacy preserving mixed collaboration
  learning},
\newblock in: \bibinfo{booktitle}{International Conference on Service-Oriented
  System Engineering (SOSE)}, \bibinfo{year}{2019}.
\bibitem[{Huang et~al.(2020)Huang, Song, Li, and Arora}]{huang2020instahide}
\bibinfo{author}{Y.~Huang}, \bibinfo{author}{Z.~Song}, \bibinfo{author}{K.~Li},
  \bibinfo{author}{S.~Arora},
\newblock \bibinfo{title}{Instahide: Instance-hiding schemes for private
  distributed learning},
\newblock in: \bibinfo{booktitle}{International Conference on Machine Learning
  (ICML)}, \bibinfo{year}{2020}.
\bibitem[{Shin et~al.(2020)Shin, Hwang, Kim, Park, Bennis, and
  Kim}]{shin2020xor}
\bibinfo{author}{M.~Shin}, \bibinfo{author}{C.~Hwang},
  \bibinfo{author}{J.~Kim}, \bibinfo{author}{J.~Park},
  \bibinfo{author}{M.~Bennis}, \bibinfo{author}{S.-L. Kim},
\newblock \bibinfo{title}{Xor mixup: Privacy-preserving data augmentation for
  one-shot federated learning},
\newblock \bibinfo{journal}{arXiv preprint arXiv:2006.05148}
  (\bibinfo{year}{2020}).
\bibitem[{Borgnia et~al.(2021)Borgnia, Geiping, Cherepanova, Fowl, Gupta,
  Ghiasi, Huang, Goldblum, and Goldstein}]{borgnia2021dp}
\bibinfo{author}{E.~Borgnia}, \bibinfo{author}{J.~Geiping},
  \bibinfo{author}{V.~Cherepanova}, \bibinfo{author}{L.~Fowl},
  \bibinfo{author}{A.~Gupta}, \bibinfo{author}{A.~Ghiasi},
  \bibinfo{author}{F.~Huang}, \bibinfo{author}{M.~Goldblum},
  \bibinfo{author}{T.~Goldstein},
\newblock \bibinfo{title}{Dp-instahide: Provably defusing poisoning and
  backdoor attacks with differentially private data augmentations},
\newblock \bibinfo{journal}{arXiv preprint arXiv:2103.02079}
  (\bibinfo{year}{2021}).
\bibitem[{Kim et~al.(2019)Kim, Kang, Pulli, and Choi}]{kim2019training}
\bibinfo{author}{T.-h. Kim}, \bibinfo{author}{D.~Kang},
  \bibinfo{author}{K.~Pulli}, \bibinfo{author}{J.~Choi},
\newblock \bibinfo{title}{Training with the invisibles: Obfuscating images to
  share safely for learning visual recognition models},
\newblock \bibinfo{journal}{arXiv preprint arXiv:1901.00098}
  (\bibinfo{year}{2019}).
\bibitem[{Xu et~al.(2019)Xu, Ren, Zhang, Zhang, Qin, and
  Ren}]{xu2019ganobfuscator}
\bibinfo{author}{C.~Xu}, \bibinfo{author}{J.~Ren}, \bibinfo{author}{D.~Zhang},
  \bibinfo{author}{Y.~Zhang}, \bibinfo{author}{Z.~Qin},
  \bibinfo{author}{K.~Ren},
\newblock \bibinfo{title}{Ganobfuscator: Mitigating information leakage under
  gan via differential privacy},
\newblock \bibinfo{journal}{Transactions on Information Forensics and Security}
   (\bibinfo{year}{2019}).
\bibitem[{Donahue and Simonyan(2019)}]{donahue2019large-arl}
\bibinfo{author}{J.~Donahue}, \bibinfo{author}{K.~Simonyan},
\newblock \bibinfo{title}{Large scale adversarial representation learning},
\newblock \bibinfo{journal}{Advances in Neural Information Processing Systems}
  \bibinfo{volume}{32} (\bibinfo{year}{2019}).
\bibitem[{Lyu et~al.(2020)Lyu, Yu, and Yang}]{lyu2020threats}
\bibinfo{author}{L.~Lyu}, \bibinfo{author}{H.~Yu}, \bibinfo{author}{Q.~Yang},
\newblock \bibinfo{title}{Threats to federated learning: A survey},
\newblock \bibinfo{journal}{arXiv preprint arXiv:2003.02133}
  (\bibinfo{year}{2020}).
\bibitem[{Pasquini et~al.(2021)Pasquini, Ateniese, and
  Bernaschi}]{pasquini2021unleashing}
\bibinfo{author}{D.~Pasquini}, \bibinfo{author}{G.~Ateniese},
  \bibinfo{author}{M.~Bernaschi},
\newblock \bibinfo{title}{Unleashing the tiger: Inference attacks on split
  learning},
\newblock in: \bibinfo{booktitle}{ACM SIGSAC Conference on Computer and
  Communications Security}, \bibinfo{year}{2021}.
\bibitem[{Li et~al.(2021)Li, Sun, Yang, Gao, Zhang, Xie, Smith, and
  Wang}]{li2021label}
\bibinfo{author}{O.~Li}, \bibinfo{author}{J.~Sun}, \bibinfo{author}{X.~Yang},
  \bibinfo{author}{W.~Gao}, \bibinfo{author}{H.~Zhang},
  \bibinfo{author}{J.~Xie}, \bibinfo{author}{V.~Smith},
  \bibinfo{author}{C.~Wang},
\newblock \bibinfo{title}{Label leakage and protection in two-party split
  learning},
\newblock \bibinfo{journal}{arXiv preprint arXiv:2102.08504}
  (\bibinfo{year}{2021}).
\bibitem[{Bertran et~al.(2019)Bertran, Martinez, Papadaki, Qiu, Rodrigues,
  Reeves, and Sapiro}]{bertran2019adversarially}
\bibinfo{author}{M.~Bertran}, \bibinfo{author}{N.~Martinez},
  \bibinfo{author}{A.~Papadaki}, \bibinfo{author}{Q.~Qiu},
  \bibinfo{author}{M.~Rodrigues}, \bibinfo{author}{G.~Reeves},
  \bibinfo{author}{G.~Sapiro},
\newblock \bibinfo{title}{Adversarially learned representations for information
  obfuscation and inference},
\newblock in: \bibinfo{booktitle}{International Conference on Machine Learning
  (ICML)}, \bibinfo{year}{2019}.
\bibitem[{Singh et~al.(2021)Singh, Chopra, Garza, Zhang, Vepakomma, Sharma, and
  Raskar}]{singh2021disco}
\bibinfo{author}{A.~Singh}, \bibinfo{author}{A.~Chopra},
  \bibinfo{author}{E.~Garza}, \bibinfo{author}{E.~Zhang},
  \bibinfo{author}{P.~Vepakomma}, \bibinfo{author}{V.~Sharma},
  \bibinfo{author}{R.~Raskar},
\newblock \bibinfo{title}{Disco: Dynamic and invariant sensitive channel
  obfuscation for deep neural networks},
\newblock in: \bibinfo{booktitle}{Conference on Computer Vision and Pattern
  Recognition (CVPR)}, \bibinfo{year}{2021}.
\bibitem[{Roy and Boddeti(2019)}]{roy2019mitigating}
\bibinfo{author}{P.~C. Roy}, \bibinfo{author}{V.~N. Boddeti},
\newblock \bibinfo{title}{Mitigating information leakage in image
  representations: A maximum entropy approach},
\newblock in: \bibinfo{booktitle}{Proceedings of the IEEE/CVF Conference on
  Computer Vision and Pattern Recognition}, \bibinfo{year}{2019}, pp.
  \bibinfo{pages}{2586--2594}.
\bibitem[{Li et~al.(2021)Li, Guo, Yang, Salim, and Chen}]{li2021deepobfuscator}
\bibinfo{author}{A.~Li}, \bibinfo{author}{J.~Guo}, \bibinfo{author}{H.~Yang},
  \bibinfo{author}{F.~D. Salim}, \bibinfo{author}{Y.~Chen},
\newblock \bibinfo{title}{Deepobfuscator: Obfuscating intermediate
  representations with privacy-preserving adversarial learning on smartphones},
\newblock in: \bibinfo{booktitle}{International Conference on
  Internet-of-Things Design and Implementation}, \bibinfo{year}{2021}.
\bibitem[{Edwards and Storkey(2016)}]{edwards2016censoring}
\bibinfo{author}{H.~Edwards}, \bibinfo{author}{A.~Storkey},
\newblock \bibinfo{title}{Censoring representations with an adversary},
\newblock in: \bibinfo{booktitle}{International Conference on Learning
  Representations (ICLR)}, \bibinfo{year}{2016}.
\bibitem[{Louizos et~al.(2016)Louizos, Swersky, Li, Welling, and
  Zemel}]{louizos2016variational}
\bibinfo{author}{C.~Louizos}, \bibinfo{author}{K.~Swersky},
  \bibinfo{author}{Y.~Li}, \bibinfo{author}{M.~Welling},
  \bibinfo{author}{R.~Zemel},
\newblock \bibinfo{title}{The variational fair autoencoder}
  (\bibinfo{year}{2016}).
\bibitem[{Madras et~al.(2018)Madras, Creager, Pitassi, and
  Zemel}]{madras2018learning}
\bibinfo{author}{D.~Madras}, \bibinfo{author}{E.~Creager},
  \bibinfo{author}{T.~Pitassi}, \bibinfo{author}{R.~Zemel},
\newblock \bibinfo{title}{Learning adversarially fair and transferable
  representations},
\newblock in: \bibinfo{booktitle}{International Conference on Machine Learning
  (ICML)}, \bibinfo{year}{2018}.
\bibitem[{Vepakomma et~al.(2020)Vepakomma, Singh, Gupta, and
  Raskar}]{vepakomma2020nopeek}
\bibinfo{author}{P.~Vepakomma}, \bibinfo{author}{A.~Singh},
  \bibinfo{author}{O.~Gupta}, \bibinfo{author}{R.~Raskar},
\newblock \bibinfo{title}{Nopeek: Information leakage reduction to share
  activations in distributed deep learning},
\newblock in: \bibinfo{booktitle}{2020 International Conference on Data Mining
  Workshops (ICDMW)}, \bibinfo{year}{2020}.
\bibitem[{Osia et~al.(2020)Osia, Shamsabadi, Sajadmanesh, Taheri, Katevas,
  Rabiee, Lane, and Haddadi}]{osia2020hybrid}
\bibinfo{author}{S.~A. Osia}, \bibinfo{author}{A.~S. Shamsabadi},
  \bibinfo{author}{S.~Sajadmanesh}, \bibinfo{author}{A.~Taheri},
  \bibinfo{author}{K.~Katevas}, \bibinfo{author}{H.~R. Rabiee},
  \bibinfo{author}{N.~D. Lane}, \bibinfo{author}{H.~Haddadi},
\newblock \bibinfo{title}{A hybrid deep learning architecture for
  privacy-preserving mobile analytics},
\newblock \bibinfo{journal}{IEEE Internet of Things Journal}
  (\bibinfo{year}{2020}).
\bibitem[{Pittaluga et~al.(2019)Pittaluga, Koppal, and
  Chakrabarti}]{pittaluga2019learning}
\bibinfo{author}{F.~Pittaluga}, \bibinfo{author}{S.~Koppal},
  \bibinfo{author}{A.~Chakrabarti},
\newblock \bibinfo{title}{Learning privacy preserving encodings through
  adversarial training},
\newblock in: \bibinfo{booktitle}{Winter Conference on Applications of Computer
  Vision (WACV)}, \bibinfo{year}{2019}.
\bibitem[{Liu et~al.(2019)Liu, Du, Shrivastava, and Zhong}]{liu2019privacy}
\bibinfo{author}{S.~Liu}, \bibinfo{author}{J.~Du},
  \bibinfo{author}{A.~Shrivastava}, \bibinfo{author}{L.~Zhong},
\newblock \bibinfo{title}{Privacy adversarial network: representation learning
  for mobile data privacy},
\newblock \bibinfo{journal}{ACM on Interactive, Mobile, Wearable and Ubiquitous
  Technologies}  (\bibinfo{year}{2019}).
\bibitem[{Wu et~al.(2018)Wu, Wang, Wang, and Jin}]{wu2018towards}
\bibinfo{author}{Z.~Wu}, \bibinfo{author}{Z.~Wang}, \bibinfo{author}{Z.~Wang},
  \bibinfo{author}{H.~Jin},
\newblock \bibinfo{title}{Towards privacy-preserving visual recognition via
  adversarial training: A pilot study},
\newblock in: \bibinfo{booktitle}{European Conference on Computer Vision
  (ECCV)}, \bibinfo{year}{2018}.
\bibitem[{Sadeghi et~al.(2019)Sadeghi, Yu, and Boddeti}]{sadeghi2019global}
\bibinfo{author}{B.~Sadeghi}, \bibinfo{author}{R.~Yu},
  \bibinfo{author}{V.~Boddeti},
\newblock \bibinfo{title}{On the global optima of kernelized adversarial
  representation learning},
\newblock in: \bibinfo{booktitle}{International Conference on Computer Vision
  (ICCV)}, \bibinfo{year}{2019}.
\bibitem[{Shokri et~al.(2017)Shokri, Stronati, Song, and
  Shmatikov}]{shokri2017membership}
\bibinfo{author}{R.~Shokri}, \bibinfo{author}{M.~Stronati},
  \bibinfo{author}{C.~Song}, \bibinfo{author}{V.~Shmatikov},
\newblock \bibinfo{title}{Membership inference attacks against machine learning
  models},
\newblock in: \bibinfo{booktitle}{Symposium on security and privacy (SP)},
  \bibinfo{year}{2017}.
\bibitem[{Fredrikson et~al.(2015)Fredrikson, Jha, and
  Ristenpart}]{fredrikson2015model}
\bibinfo{author}{M.~Fredrikson}, \bibinfo{author}{S.~Jha},
  \bibinfo{author}{T.~Ristenpart},
\newblock \bibinfo{title}{Model inversion attacks that exploit confidence
  information and basic countermeasures},
\newblock in: \bibinfo{booktitle}{ACM SIGSAC conference on computer and
  communications security}, \bibinfo{year}{2015}.
\bibitem[{Yin et~al.(2019)Yin, Lopes, Shlens, Cubuk, and
  Gilmer}]{yin2019fourier}
\bibinfo{author}{D.~Yin}, \bibinfo{author}{R.~G. Lopes},
  \bibinfo{author}{J.~Shlens}, \bibinfo{author}{E.~D. Cubuk},
  \bibinfo{author}{J.~Gilmer},
\newblock \bibinfo{title}{A fourier perspective on model robustness in computer
  vision},
\newblock in: \bibinfo{booktitle}{Advances in neural information processing
  systems (NeurIPS)}, \bibinfo{year}{2019}.
\bibitem[{Wang et~al.(2020)Wang, Wu, Huang, and Xing}]{wang2020high}
\bibinfo{author}{H.~Wang}, \bibinfo{author}{X.~Wu}, \bibinfo{author}{Z.~Huang},
  \bibinfo{author}{E.~P. Xing},
\newblock \bibinfo{title}{High-frequency component helps explain the
  generalization of convolutional neural networks},
\newblock in: \bibinfo{booktitle}{Conference on Computer Vision and Pattern
  Recognition (CVPR)}, \bibinfo{year}{2020}.
\bibitem[{Geirhos et~al.(2019)Geirhos, Rubisch, Michaelis, Bethge, Wichmann,
  and Brendel}]{geirhos2018imagenet}
\bibinfo{author}{R.~Geirhos}, \bibinfo{author}{P.~Rubisch},
  \bibinfo{author}{C.~Michaelis}, \bibinfo{author}{M.~Bethge},
  \bibinfo{author}{F.~A. Wichmann}, \bibinfo{author}{W.~Brendel},
\newblock \bibinfo{title}{Imagenet-trained cnns are biased towards texture;
  increasing shape bias improves accuracy and robustness.},
\newblock in: \bibinfo{booktitle}{International Conference on Learning
  Representations (ICLR)}, \bibinfo{year}{2019}.
\bibitem[{Abello et~al.(2021)Abello, Hirata, and Wang}]{abello2021dissecting}
\bibinfo{author}{A.~A. Abello}, \bibinfo{author}{R.~Hirata},
  \bibinfo{author}{Z.~Wang},
\newblock \bibinfo{title}{Dissecting the high-frequency bias in convolutional
  neural networks},
\newblock in: \bibinfo{booktitle}{Proceedings of the IEEE/CVF Conference on
  Computer Vision and Pattern Recognition}, \bibinfo{year}{2021}, pp.
  \bibinfo{pages}{863--871}.
\bibitem[{Ilyas et~al.(2019)Ilyas, Santurkar, Tsipras, Engstrom, Tran, and
  Madry}]{ilyas2019adversarial}
\bibinfo{author}{A.~Ilyas}, \bibinfo{author}{S.~Santurkar},
  \bibinfo{author}{D.~Tsipras}, \bibinfo{author}{L.~Engstrom},
  \bibinfo{author}{B.~Tran}, \bibinfo{author}{A.~Madry},
\newblock \bibinfo{title}{Adversarial examples are not bugs, they are
  features},
\newblock \bibinfo{journal}{Advances in neural information processing systems
  (NeurIPS)}  (\bibinfo{year}{2019}).
\bibitem[{Lim(1990)}]{lim1990two}
\bibinfo{author}{J.~S. Lim},
\newblock \bibinfo{title}{Two-dimensional signal and image processing},
\newblock \bibinfo{journal}{Englewood Cliffs}  (\bibinfo{year}{1990}).
\bibitem[{Ronneberger et~al.(2015)Ronneberger, Fischer, and
  Brox}]{ronneberger2015u}
\bibinfo{author}{O.~Ronneberger}, \bibinfo{author}{P.~Fischer},
  \bibinfo{author}{T.~Brox},
\newblock \bibinfo{title}{U-net: Convolutional networks for biomedical image
  segmentation},
\newblock in: \bibinfo{booktitle}{International Conference on Medical image
  computing and computer-assisted intervention}, \bibinfo{year}{2015}.
\bibitem[{He et~al.(2016)He, Zhang, Ren, and Sun}]{he2016deep}
\bibinfo{author}{K.~He}, \bibinfo{author}{X.~Zhang}, \bibinfo{author}{S.~Ren},
  \bibinfo{author}{J.~Sun},
\newblock \bibinfo{title}{Deep residual learning for image recognition},
\newblock in: \bibinfo{booktitle}{Conference on computer vision and pattern
  recognition (CVPR)}, \bibinfo{year}{2016}.
\bibitem[{Liu et~al.(2015)Liu, Luo, Wang, and Tang}]{liu2015faceattributes}
\bibinfo{author}{Z.~Liu}, \bibinfo{author}{P.~Luo}, \bibinfo{author}{X.~Wang},
  \bibinfo{author}{X.~Tang},
\newblock \bibinfo{title}{Deep learning face attributes in the wild},
\newblock in: \bibinfo{booktitle}{International Conference on Computer Vision
  (ICCV)}, \bibinfo{year}{2015}.
\bibitem[{Karkkainen and Joo(2021)}]{karkkainen2021fairface}
\bibinfo{author}{K.~Karkkainen}, \bibinfo{author}{J.~Joo},
\newblock \bibinfo{title}{Fairface: Face attribute dataset for balanced race,
  gender, and age for bias measurement and mitigation},
\newblock in: \bibinfo{booktitle}{Winter Conference on Applications of Computer
  Vision (WACV)}, \bibinfo{year}{2021}.
\bibitem[{Krizhevsky(2009)}]{Krizhevsky09learningmultiple}
\bibinfo{author}{A.~Krizhevsky}, \bibinfo{title}{Learning multiple layers of
  features from tiny images}, \bibinfo{type}{Technical Report},
  \bibinfo{year}{2009}.
\bibitem[{Kingma and Ba(2014)}]{kingma2014adam}
\bibinfo{author}{D.~P. Kingma}, \bibinfo{author}{J.~Ba},
\newblock \bibinfo{title}{Adam: A method for stochastic optimization},
\newblock \bibinfo{journal}{arXiv preprint arXiv:1412.6980}
  (\bibinfo{year}{2014}).
\bibitem[{Wang et~al.(2004)Wang, Bovik, Sheikh, and Simoncelli}]{zhou2004ssim}
\bibinfo{author}{Z.~Wang}, \bibinfo{author}{A.~Bovik},
  \bibinfo{author}{H.~Sheikh}, \bibinfo{author}{E.~Simoncelli},
\newblock \bibinfo{title}{Image quality assessment: from error visibility to
  structural similarity},
\newblock \bibinfo{journal}{Transactions on Image Processing}
  (\bibinfo{year}{2004}).
\bibitem[{Wang et~al.(2003)Wang, Simoncelli, and
  Bovik}]{wang2003multiscale-msssim}
\bibinfo{author}{Z.~Wang}, \bibinfo{author}{E.~P. Simoncelli},
  \bibinfo{author}{A.~C. Bovik},
\newblock \bibinfo{title}{Multiscale structural similarity for image quality
  assessment},
\newblock in: \bibinfo{booktitle}{The Thrity-Seventh Asilomar Conference on
  Signals, Systems \& Computers, 2003}, volume~\bibinfo{volume}{2},
  \bibinfo{organization}{Ieee}, \bibinfo{year}{2003}, pp.
  \bibinfo{pages}{1398--1402}.
\bibitem[{Horé and Ziou(2010)}]{hore2010psnr}
\bibinfo{author}{A.~Horé}, \bibinfo{author}{D.~Ziou},
\newblock \bibinfo{title}{Image quality metrics: Psnr vs. ssim},
\newblock in: \bibinfo{booktitle}{International Conference on Pattern
  Recognition}, \bibinfo{year}{2010}.
\bibitem[{Zhang et~al.(2018)Zhang, Isola, Efros, Shechtman, and
  Wang}]{zhang2018unreasonable-lpips}
\bibinfo{author}{R.~Zhang}, \bibinfo{author}{P.~Isola}, \bibinfo{author}{A.~A.
  Efros}, \bibinfo{author}{E.~Shechtman}, \bibinfo{author}{O.~Wang},
\newblock \bibinfo{title}{The unreasonable effectiveness of deep features as a
  perceptual metric},
\newblock in: \bibinfo{booktitle}{Proceedings of the IEEE conference on
  computer vision and pattern recognition}, \bibinfo{year}{2018}, pp.
  \bibinfo{pages}{586--595}.
\bibitem[{Karras et~al.(2020)Karras, Laine, Aittala, Hellsten, Lehtinen, and
  Aila}]{Karras_2020_CVPR}
\bibinfo{author}{T.~Karras}, \bibinfo{author}{S.~Laine},
  \bibinfo{author}{M.~Aittala}, \bibinfo{author}{J.~Hellsten},
  \bibinfo{author}{J.~Lehtinen}, \bibinfo{author}{T.~Aila},
\newblock \bibinfo{title}{Analyzing and improving the image quality of
  stylegan},
\newblock in: \bibinfo{booktitle}{Proceedings of the IEEE/CVF Conference on
  Computer Vision and Pattern Recognition (CVPR)}, \bibinfo{year}{2020}.
\bibitem[{Howard et~al.(2019)Howard, Sandler, Chu, Chen, Chen, Tan, Wang, Zhu,
  Pang, Vasudevan et~al.}]{howard2019searching-mobilenetv3}
\bibinfo{author}{A.~Howard}, \bibinfo{author}{M.~Sandler},
  \bibinfo{author}{G.~Chu}, \bibinfo{author}{L.-C. Chen},
  \bibinfo{author}{B.~Chen}, \bibinfo{author}{M.~Tan},
  \bibinfo{author}{W.~Wang}, \bibinfo{author}{Y.~Zhu},
  \bibinfo{author}{R.~Pang}, \bibinfo{author}{V.~Vasudevan}, et~al.,
\newblock \bibinfo{title}{Searching for mobilenetv3},
\newblock in: \bibinfo{booktitle}{Proceedings of the IEEE/CVF International
  Conference on Computer Vision}, \bibinfo{year}{2019}.

\end{thebibliography}

\end{document}